\documentclass[final]{cvpr}

\usepackage{times}
\usepackage{epsfig}
\usepackage{graphicx}
\usepackage{amsmath}
\usepackage{amssymb}

\usepackage{algorithm}
\usepackage{algorithmic}
\usepackage{bm}
\usepackage{mathtools}
\usepackage{makecell}
\usepackage{enumitem}
\usepackage{caption}
\usepackage{lipsum}
\usepackage{cuted}
\usepackage{siunitx}
\usepackage{booktabs}

\usepackage{subcaption}
\usepackage{multirow}
\makeatletter
\DeclareRobustCommand\onedot{\futurelet\@let@token\@onedot}
\def\@onedot{\ifx\@let@token.\else.\null\fi\xspace}

\def\eg{\emph{e.g}\onedot} 
\def\ie{\emph{i.e}\onedot} 
 
\def\etc{\emph{etc}\onedot} \def\vs{\emph{vs}\onedot}

\makeatother

\usepackage[pagebackref=true,breaklinks=true,colorlinks,bookmarks=false]{hyperref}

\def\cvprPaperID{7718} 
\def\confYear{CVPR 2021}

\begin{document}

\title{Diverse Branch Block: Building a Convolution as an Inception-like Unit}

\author{Xiaohan Ding \textsuperscript{1}\thanks{This work is supported by The National Key Research and Development Program of China (No. 2017YFA0700800), the National Natural Science Foundation of China (No.61925107, No.U1936202) and Beijing Academy of Artificial Intelligence (BAAI). Xiaohan Ding is funded by the Baidu Scholarship Program 2019 (\url{http://scholarship.baidu.com/}). This work is done during Xiaohan Ding's internship at MEGVII.} \quad Xiangyu Zhang \textsuperscript{2} \quad Jungong Han \textsuperscript{3} \quad Guiguang Ding \textsuperscript{1}\thanks{Corresponding author.} \\
	\textsuperscript{1} Beijing National Research Center for Information Science and Technology (BNRist); \\School of Software, Tsinghua University, Beijing, China \\
	\textsuperscript{2} MEGVII Technology \\
	\textsuperscript{3} Computer Science Department, Aberystwyth University, SY23 3FL, UK \\
	\tt\small dxh17@mails.tsinghua.edu.cn \quad zhangxiangyu@megvii.com\\
	\tt\small jungonghan77@gmail.com \quad dinggg@tsinghua.edu.cn\\
}

\maketitle

\pagestyle{empty}
\thispagestyle{empty}

\begin{abstract}
We propose a universal building block of Convolutional Neural Network (ConvNet) to improve the performance without any inference-time costs. The block is named Diverse Branch Block (DBB), which enhances the representational capacity of a single convolution by combining diverse branches of different scales and complexities to enrich the feature space, including sequences of convolutions, multi-scale convolutions, and average pooling. After training, a DBB can be equivalently converted into a single conv layer for deployment. Unlike the advancements of novel ConvNet architectures, DBB complicates the training-time microstructure while maintaining the macro architecture, so that it can be used as a drop-in replacement for regular conv layers of any architecture. In this way, the model can be trained to reach a higher level of performance and then transformed into the original inference-time structure for inference. DBB improves ConvNets on image classification (up to 1.9\% higher top-1 accuracy on ImageNet), object detection and semantic segmentation. The PyTorch code and models are released at \url{https://github.com/DingXiaoH/DiverseBranchBlock}.
\end{abstract}

\section{Introduction}
Improving the performance of Convolutional Neural Network (ConvNet) has always been a heated research topic. On one hand, the advancements in architecture design, \eg, the Inception models \cite{szegedy2015going,szegedy2016rethinking,szegedy2017inception,ioffe2015batch}, have revealed that the multi-branch topology and combination of various paths with different scales and complexities can enrich the feature space and improve the performance. However, the complicated structure usually slows down the inference, as a combination of small operators (\eg, concatenation of $1\times1$ conv and pooling) is not friendly to the devices with strong parallel computing powers like GPU \cite{ma2018shufflenet}.

On the other hand, more parameters and connections usually lead to higher performance, but the size of ConvNet we deploy cannot increase arbitrarily due to the business requirements and hardware constraints. Considering this, we usually judge the quality of a ConvNet by the trade-off between performance and inference-time costs such as the latency, memory footprint and number of parameters. In the common cases, we train the models on powerful GPU workstations and deploy them onto efficiency-sensitive devices, so we consider it acceptable to improve the performance with the costs of more training resources, as long as the deployed model keeps the same size. 

In this paper, we seek to insert complicated structures into numerous ConvNet architectures to improve the performance while keeping the original inference-time costs. To this end, we decouple the training-time and inference-time network structure by \textit{complicating the model only during training} and converting it back into the original structure for inference. Naturally, we require such extra training-time structures to be \textbf{1)} effective in improving the training-time model's performance and \textbf{2)} able to transform into the original inference-time structure.

\begin{figure*}
	\begin{center}
		\includegraphics[width=0.9\linewidth]{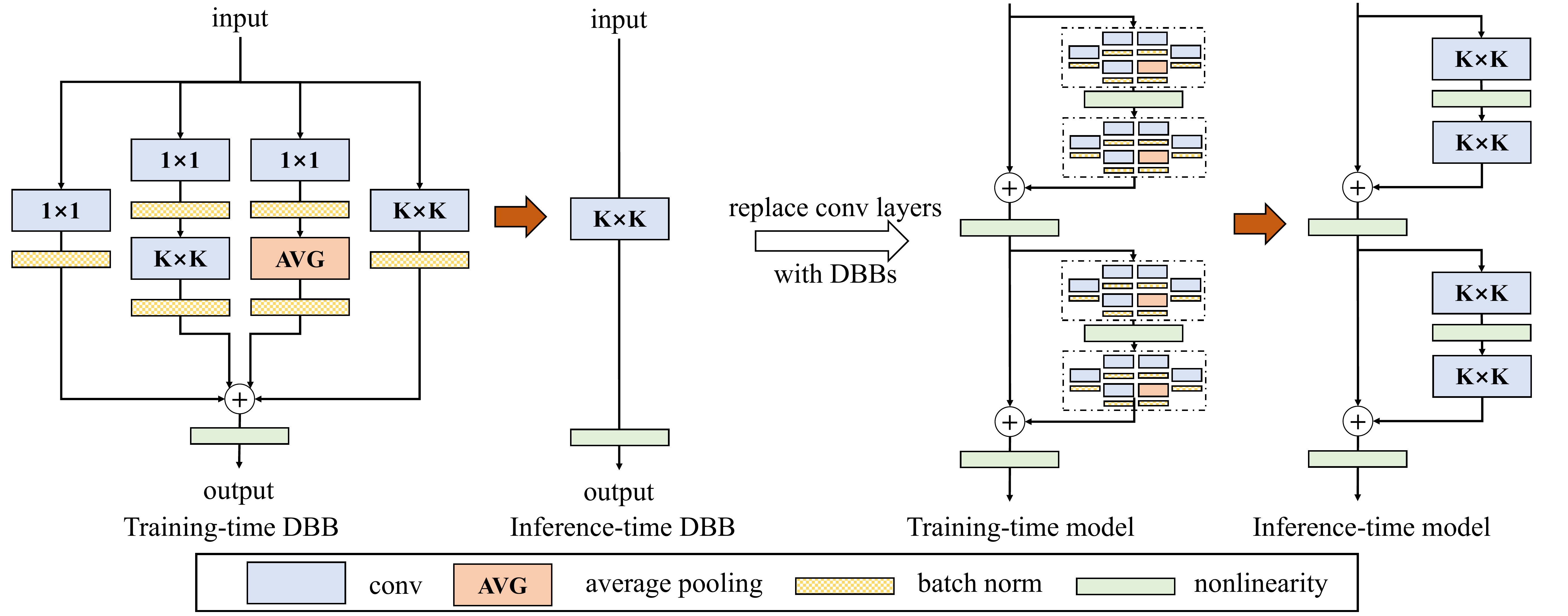}
				\vspace{-0.1in}
		\caption{A representative design of Diverse Branch Block (DBB). The block can be equivalently transformed into a regular conv layer for deployment, so that we can complicate the training-time microstructures of ConvNet without affecting the macro architecture (\eg, ResNet) or the inference-time structure. Note that it is only an instance we used, and one may utilize the six transformations summarized in this paper (Sect. \ref{sect-transforms}) to customize a DBB with more complicated structures. }
		\label{fig-topology}
	\end{center}
	\vspace{-0.1in}
\end{figure*}

For the usability and universality, we upgrade the basic ConvNet component, $K\times K$ conv, into a powerful block named Diverse Branch Block (DBB) (Fig. \ref{fig-topology}). As a building block, DBB is complementary to the other efforts to improve ConvNet, \eg, architecture design \cite{he2016deep,simonyan2014very,mobilev1,ma2018shufflenet,mbv2,zhang2018shufflenet}, neural architecture search \cite{chen2019efficient,zoph2016neural,real2019regularized,liu2018darts,liu2018progressive}, data augmentation and training methods \cite{sinha2020curriculum,cubuk2019autoaugment,zhang2017mixup} and fulfills the above two requirements by the following two properties:
\begin{itemize}[noitemsep,nolistsep,topsep=0pt,parsep=0pt,partopsep=0pt]
	\item A DBB adopts a multi-branch topology with multi-scale convolutions, sequential $1\times1$ - $K\times K$ convolutions, average pooling and branch addition. Such operations with various receptive fields and paths with different complexities can enrich the feature space, just like the Inception architectures.
	\item A DBB can be equivalently transformed into a single conv for inference. Given an architecture, we can replace some regular conv layers with DBB to build more complicated miscrostructures for training, and convert it back into the original structure so that there will be no extra inference-time costs.
\end{itemize}

More precisely, we do not derive the parameters for inference before each forwarding. Instead, we \textit{convert the model after training once for all}, then we only save and use the resultant model, and the trained model can be discarded. The idea of converting a DBB into a conv can be categorized into \textit{structural re-parameterization}, which means parameterizing a structure with the parameters transformed from another structure, together with a concurrent work \cite{ding2021repvgg}. Though a DBB and a regular conv layer have the same inference-time structure, the former has higher representational capacity. Through a series of ablation experiments, we attribute such effectiveness to the \textit{diverse connections} (paths with different scales and complexities) which resembles Inception units, and the \textit{training-time nonlinearity} brought by batch normalization \cite{ioffe2015batch}. Compared to some counterparts with duplicate paths or purely linear branches (Fig. \ref{fig-counterparts}), DBB shows better performance (Table. \ref{table-diversity}).

We summarize our contributions as follows.
\begin{itemize}[noitemsep,nolistsep,topsep=0pt,parsep=0pt,partopsep=0pt]
	\item We propose to incorporate abundant microstructures into various ConvNet architectures to improve the performance but keep the original macro architecture.	
	\item We propose DBB, a universal building block, and summarize six transformations (Fig. \ref{fig-transform}) to convert a DBB into a single convolution, so it is cost-free for the users.
	\item We present a representative Inception-like DBB instance and show that it improves the performance on ImageNet \cite{deng2009imagenet} (\eg, up to 1.9\% higher top-1 accuracy), COCO detection \cite{coco} and Cityscapes \cite{cityscapes}.
\end{itemize}

\section{Related Work}

\subsection{Multi-branch Architectures}
Inception \cite{ioffe2015batch,szegedy2017inception,szegedy2015going,szegedy2016rethinking} architectures employed multi-branch structures to enrich the feature space, which proved the significance of diverse connections, various receptive fields and the combination of multiple branches. DBB borrows the idea of using multi-branch topology, but the difference lies in that \textbf{1)} DBB is a building block that can be used on numerous architectures, and \textbf{2)} each branch of DBB can be converted into a conv, so that the combination of such branches can be merged into a single conv, which is much faster than a real Inception unit. We will show the superiority of diverse branches over duplicate ones (Table. \ref{table-diversity}), and the most interesting discovery is that combining two branches with different representational capacity (\eg, a $1\times1$ conv and a $3\times3$ conv) is better than two strong-capacity branches (\eg, two $3\times3$ convolutions), which may in turn shed light on ConvNet architecture design. 

\subsection{ConvNet Components for Better Performance}
There have been some novel components to improve ConvNets. For examples, Squeeze-and-Excitation (SE) block \cite{seblock} and Efficient Convolutional Attention (ECA) block \cite{ecanet} utilizes the attention mechanism to recalibrate the features, Octave Convolution \cite{octave} reduces the spatial redundancy of regular convolution, Deformable Convolution \cite{deformable} augments the spatial sampling locations with learnable offsets, dilated convolution expands the receptive field \cite{yu2015multi}, BlurPool \cite{zhang2019making} brings back the shift-invariance, \etc. DBB is complementary to these components because it only upgrades a fundamental building block: the conv layer.

\subsection{Structural Re-parameterization}

This paper and a concurrent work, RepVGG \cite{ding2021repvgg}, are the first to use structural re-parameterization to term the methodology that parameterizes a structure with the parameters transformed from another structure. ExpandNet \cite{guo2020expandnets}, DO-Conv \cite{cao2020conv} and ACNet \cite{ding2019acnet} can also be categorized into structural re-parameterization in the sense that they convert a block into a conv. For example, ACNet uses Asymmetric Convolution Block (ACB, as shown in Fig. \ref{fig-counterparts}d) to strengthen the skeleton of conv kernel (\ie, the crisscross part). Compared to DBB, it is also designed for improving ConvNet without extra inference-time costs. However, the difference is that ACNet was motivated by an observation that the parameters of the skeleton were larger in magnitude and thus sought to make them even larger, whereas we focus on a different perspective. We found out that average pooling, $1\times 1$ conv, and $1\times1$ - $K\times K$ sequential conv are more effective, as they provide paths with different complexities, and allow the usage of more training-time nonlinearity. Besides, ACB can be viewed as a special case of DBB, since the $1\times K$ and $K\times 1$ conv layers can be augmented to $K\times K$ via Transform VI (Fig. \ref{fig-transform}) and merged into the square kernel via Transform II.

\subsection{Other ConvNet Re-parameterization Methods}

Some works can be referred to as re-parameterization, but not \textit{structural} re-parameterization. For examples, a recent NAS \cite{liu2018darts,zoph2016neural} method \cite{chen2019efficient} used meta-kernels to re-parameterize a kernel and supplemented the widths and heights of such meta-kernels into the search space. Soft Conditional Computation (SCC) \cite{yang2019soft} or CondConv \cite{condconv} can be viewed as data-dependent kernel re-parameterization, as it generated the weights for multiple kernels of the same shape, then derived a kernel as the weighted sum of all such kernels to participate in the convolution. Note that SCC introduced a significant number of parameters into the deployed model. These re-parameterization methods differ from ACB and DBB in that the former ``re-param'' means deriving a set of new parameters with some meta parameters (\eg, the meta kernels \cite{chen2019efficient}) then using the new parameters for the other computations, while the latter means converting the parameters of a trained model to parameterize another one.

\section{Diverse Branch Block}

\subsection{The Linearity of Convolution}
The parameters of a conv layer with $C$ input channels, $D$ output channels and kernel size $K\times K$ reside in the conv kernel, which is a 4th-order tensor $\bm{F}\in \mathbb{R}^{D\times C\times K\times K}$, and an optional bias $\bm{b}\in\mathbb{R}^{D}$. It takes a $C$-channel feature map $\bm{I}\in\mathbb{R}^{C\times H\times W}$ as input and outputs a $D$-channel feature map $\bm{O}\in\mathbb{R}^{D\times H^\prime\times W^\prime}$, where $H^\prime$ and $W^\prime$ are determined by $K$, padding and stride configurations. We use $\circledast$ to denote the convolution operator, and formulate the bias-adding as replicating the bias $\bm{b}$ into $\text{REP}(\bm{b})\in\mathbb{R}^{D\times H^\prime\times W^\prime}$ and adding it onto the results of convolution. Formally,
\begin{equation}\label{eq-conv}
\bm{O}=\bm{I}\circledast\bm{F} + \text{REP}(\bm{b}) \,. 
\end{equation}
The value at $(h, w)$ on the $j$-th output channel is given by
\begin{equation}\label{eq-slid-window}
\bm{O}_{j, h, w} = \sum_{c=1}^{C}\sum_{u=1}^{K}\sum_{v=1}^{K} \bm{F}_{j,c,u,v} \bm{X}(c,h,w)_{u,v} + \bm{b}_j \,,
\end{equation}
where $\bm{X}(c,h,w)\in\mathbb{R}^{K\times K}$ is the sliding window on the $c$-th channel of $\bm{I}$ corresponding to the position $(h, w)$ on $\bm{O}$. Such a correspondence is determined by the padding and stride. The linearity of conv can be easily derived from Eq.~\ref{eq-slid-window}, which includes the \textit{homogeneity} and \textit{additivity}, 
\begin{equation}
\bm{I}\circledast(p\bm{F})=p(\bm{I}\circledast\bm{F}) \,, \forall p \in \mathbb{R} \,,
\end{equation}
\begin{equation}
\bm{I}\circledast\bm{F}^{(1)}+\bm{I}\circledast\bm{F}^{(2)}=\bm{I}\circledast(\bm{F}^{(1)}+\bm{F}^{(2)}) \,.
\end{equation}
Note that the additivity holds only if the two convolutions have the same configurations (\eg, number of channels, kernel size, stride, padding, \etc), so that they share the same sliding window correspondence $X$.

\subsection{A Convolution for Diverse Branches}\label{sect-transforms}
\begin{figure*}
	\begin{center}
		\includegraphics[width=0.75\linewidth]{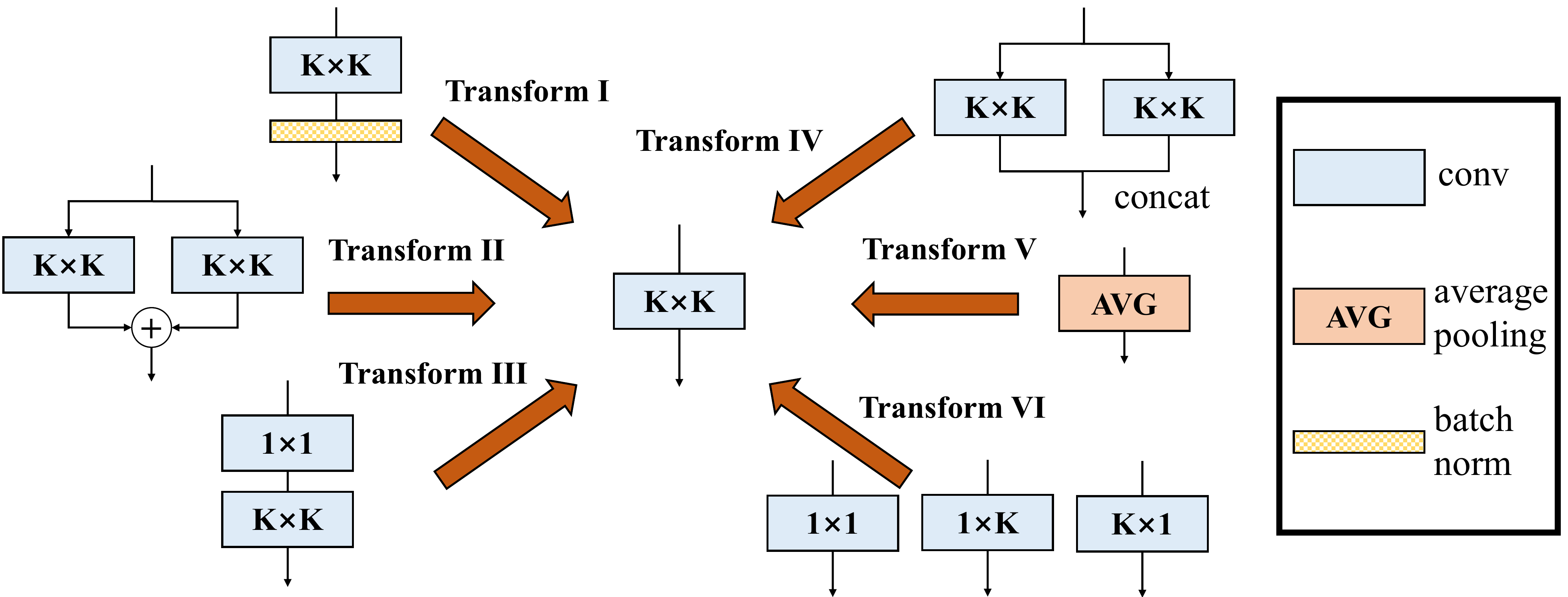}
		\vspace{-0.1in}
		\caption{Six transformations we use to implement an inference-time DBB by a regular convolutional layer.}
		\label{fig-transform}
	\end{center}
\vspace{-0.2in}
\end{figure*}

In this subsection, we summarize six transformations (Fig. \ref{fig-transform}) to transform a DBB with batch normalization (BN), branch addition, depth concatenation, multi-scale operations, average pooling and sequences of convolutions. 

\paragraph{Transform I: a conv for conv-BN}
We usually equip a conv with a BN layer, which performs channel-wise normalization and linear scaling. Let $j$ be the channel index, $\bm{\mu}_j$ and $\bm{\sigma}_j$ be the accumulated channel-wise mean and standard deviation, $\bm{\gamma}_j$ and $\bm{\beta}_j$ be the learned scaling factor and bias term, respectively, the output channel $j$ becomes
\begin{equation}\label{eq-conv-with-bn}
\bm{O}_{j,:,:}= ((\bm{I}\circledast\bm{F})_{j,:,:} - \bm{\mu}_j)\frac{\bm{\gamma}_j}{\bm{\sigma}_j} + \bm{\beta}_j \,.
\end{equation}

The homogeneity of conv enables to fuse BN into the preceding conv for inference. In practice, we simply build a single conv with kernel $\bm{F}^\prime$ and bias $\bm{b}^\prime$, assign the values transformed from the parameters of the original conv-BN sequence, then save the model for inference. By Eq. \ref{eq-conv}, \ref{eq-conv-with-bn}, we construct $\bm{F}^\prime$ and $\bm{b}^\prime$ for every output channel $j$ as
\begin{equation}\label{eq-bn-fusion}
\bm{F}^\prime_{j,:,:,:} \gets \frac{\bm{\gamma}_j}{\bm{\sigma}_j}\bm{F}_{j,:,:,:} \,,
\quad\quad
\bm{b}^\prime_j \gets -\frac{\bm{\mu}_j \bm{\gamma}_j}{\bm{\sigma}_j} + \bm{\beta}_j \,.
\end{equation}

\paragraph{Transform II: a conv for branch addition}
The additivity ensures that if the outputs of two or more conv layers with the same configurations are added up, we can merge them into a single conv. For a conv-BN, we should perform Transform I first. Obviously, we merge two convolutions by
\begin{equation}
\bm{F}^\prime \gets \bm{F}^{(1)} + \bm{F}^{(2)} \,,
\quad\quad
\bm{b}^\prime \gets \bm{b}^{(1)} + \bm{b}^{(2)} \,.
\end{equation}

The above formulas only apply to conv layers with the same configurations. Though merging such branches can strengthen the model to some extent (Table. \ref{table-diversity}), we wish to combine diverse branches to further improve the performance. In the following, we introduce some forms of branches that can be equivalently transformed into a single conv. After constructing $K\times K$ conv for each branch via multiple transformations, we use Transform II to merge all such branches into one single conv.

\paragraph{Transform III: a conv for sequential convolutions}
We can merge a sequence of $1\times1$ conv - BN - $K\times K$ conv - BN into one single $K\times K$ conv. We temporarily assume the conv is dense (\ie, number of groups $g=1$). The groupwise case with $g>1$ will be realized by Transform IV. We assume the kernel shapes of the $1\times1$ and $K\times K$ layers are $D\times C\times 1\times 1$ and $E\times D\times K\times K$, respectively, where $D$ can be arbitrary. We first fuse the two BN layers into the two conv layers to obtain $\bm{F}^{(1)} \in \mathbb{R}^{D\times C\times 1\times 1}$, $\bm{b}^{(1)} \in \mathbb{R}^{D}$, $\bm{F}^{(2)} \in \mathbb{R}^{E\times D\times K\times K}$, and $\bm{b}^{(2)} \in \mathbb{R}^{E}$. The output is
\begin{equation}\label{eq-output-sequence}
\bm{O}^{\prime} = (\bm{I}\circledast\bm{F}^{(1)} + \text{REP}(\bm{b}^{(1)}))\circledast\bm{F}^{(2)} + \text{REP}(\bm{b}^{(2)}) \,.
\end{equation}
We desire the expressions of the kernel and bias of a single conv, $\bm{F}^\prime$ and $\bm{b}^\prime$, which satisfies
\begin{equation}
\bm{O}^{\prime} = \bm{I}\circledast\bm{F}^\prime + \text{REP}(\bm{b}^\prime) \,.
\end{equation}

Applying the additivity of conv to Eq. \ref{eq-output-sequence}, we have
\begin{equation}\label{eq-sequence-additivity}
\bm{O}^\prime = \bm{I}\circledast\bm{F}^{(1)}\circledast\bm{F}^{(2)} + \text{REP}(\bm{b}^{(1)})\circledast\bm{F}^{(2)} + \text{REP}(\bm{b}^{(2)}) \,.
\end{equation}

As $\bm{I}\circledast\bm{F}^{(1)}$ is $1\times 1$ conv, which performs only channel-wise linear combination but no spatial aggregation, we can merge it into the $K\times K$ conv by linearly recombining the parameters in $K\times K$ kernel. It is easy to verify that such a transformation can be accomplished by transpose conv,
\begin{equation}
\bm{F}^\prime \gets \bm{F}^{(2)} \circledast \text{TRANS}(\bm{F}^{(1)}) \,,
\end{equation}
where $\text{TRANS}(\bm{F}^{(1)}) \in \mathbb{R}^{C\times D\times 1 \times 1}$ is the tensor transposed from $\bm{F}^{(1)}$. The second term of Eq. \ref{eq-sequence-additivity} is convolutions on constant matrices, so the outputs are also constant matrices. Formally, let $\bm{P}\in\mathbb{R}^{H\times W}$ be a constant matrix where every entry equals $p$, $\ast$ be the 2D conv operator, $\bm{W}$ be a 2D conv kernel, the result is a constant matrix proportional to $p$ and the sum of all the kernel elements, \ie, 
\begin{equation}
(\bm{P}\ast\bm{W})_{:,:}= p \text{SUM}(\bm{W}) \,.
\end{equation}
Based on this observation, we construct $\hat{\bm{b}}$ as
\begin{equation}
\hat{\bm{b}}_j \gets \sum_{d=1}^{D}\sum_{u=1}^{K}\sum_{v=1}^{K} \bm{b}^{(1)}_d \bm{F}^{(2)}_{j,d,u,v} \,, 1\leq j \leq E \,.
\end{equation}
Then it is easy to verify
\begin{equation}
\text{REP}(\bm{b}^{(1)})\circledast\bm{F}^{(2)} = \text{REP}(\hat{\bm{b}}) \,.
\end{equation}
Then we have 
\begin{equation}
\bm{b}^\prime \gets \hat{\bm{b}} + \bm{b}^{(2)} \,.
\end{equation}

Notably, for a $K\times K$ conv that zero-pads the input, Eq.~\ref{eq-output-sequence} does not hold because $\bm{F}^{(2)}$ does not convolve on the result of $\bm{I}\circledast\bm{F}^{(1)} + \text{REP}(\bm{b}^{(1)})$ (but an additional circle of zero pixels). The solution is to either \textbf{A)} configure the first conv with padding and the second without, or \textbf{B)} pad by $\bm{b}^{(1)}$. An efficient implementation of the latter is customizing the first BN to \textbf{1)} batch-normalize the input as usual, \textbf{2)} calculate $\bm{b}^{(1)}$ (Eq. \ref{eq-bn-fusion}), \textbf{3)} pad the batch-normalized result with $\bm{b}^{(1)}$, \ie, pad every channel $j$ with a circle of $\bm{b}^{(1)}_j$ instead of 0.

\begin{figure}
	\begin{center}
		\includegraphics[width=0.9\linewidth]{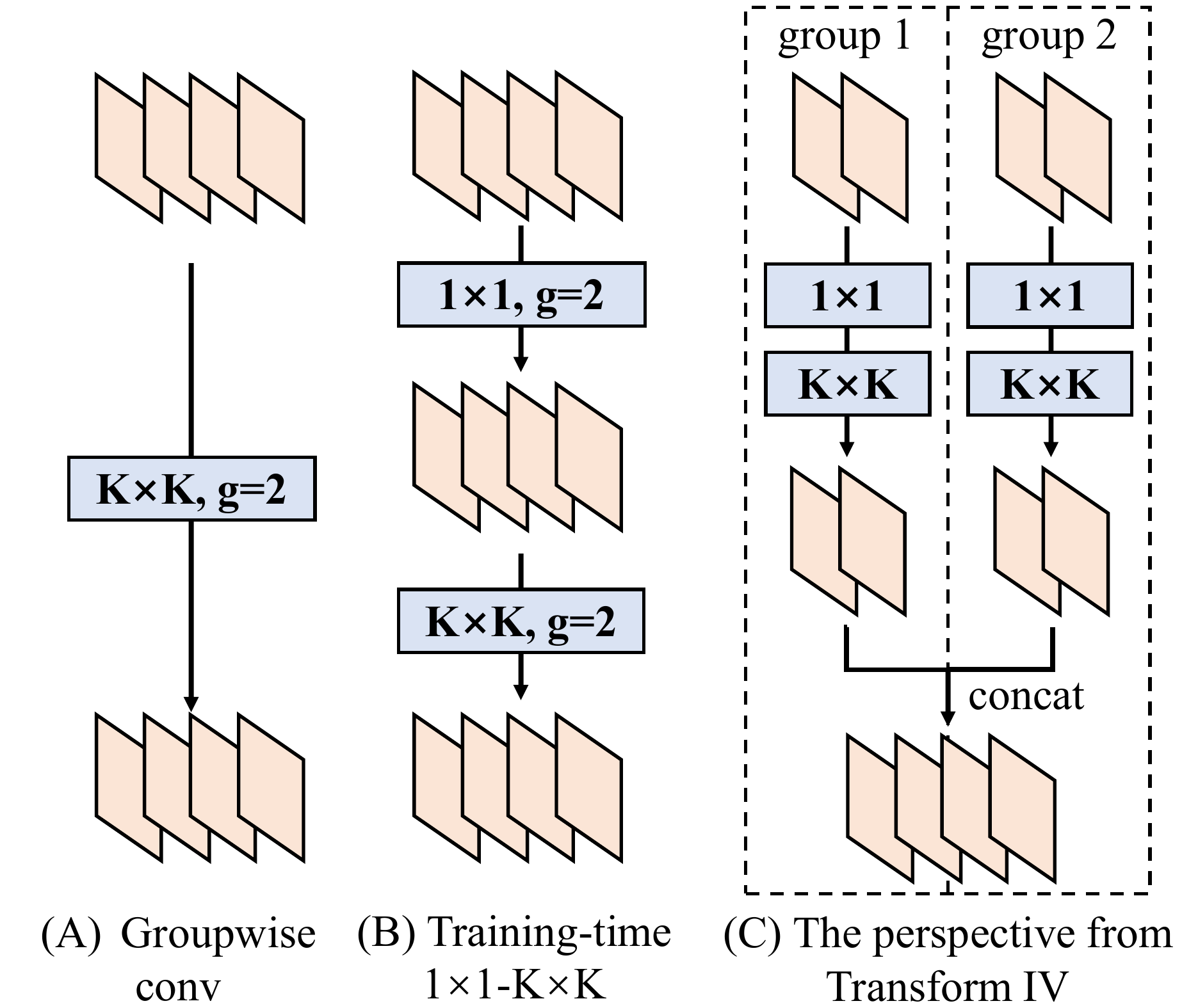}
		\caption{An example of converting $1\times1$ - $K\times K$ sequence with the number of groups $g>1$. Assume the input and output are 4-channel feature maps and $g=2$, the $1\times1$ and $K\times K$ layers should be configured with $g=2$, too. For the transformation, we split the layers into $g$ groups, perform Transform III separately, and Transform IV to concatenate the resultant kernels and biases.}
		\label{fig-dw}
	\end{center}
\vspace{-0.2in}
\end{figure}

\paragraph{Transform IV: a conv for depth concatenation}
Inception units use depth concatenation to combine branches. But when such branches each contain only one conv with the same configurations, the depth concatenation is equivalent to a conv with a kernel concatenated along the axis differentiating the output channels (\eg, the first axis in our formulation). Given $\bm{F}^{(1)}\in\mathbb{R}^{D_1\times C\times K\times K}$, $\bm{b}^{(1)}\in\mathbb{R}^{D_1}$, $\bm{F}^{(2)}\in\mathbb{R}^{D_2\times C\times K\times K}$, $\bm{b}^{(2)}\in\mathbb{R}^{D_2}$, we concatenate them into $\bm{F}^\prime \in\mathbb{R}^{(D_1 + D_2)\times C\times K\times K}$, $\bm{b}^\prime\in\mathbb{R}^{D_1 + D_2}$. Obviously,
\begin{equation}
\begin{aligned}
&\text{CONCAT}(\bm{I}\circledast\bm{F}^{(1)} + \text{REP}(\bm{b}^{(1)}), \bm{I}\circledast\bm{F}^{(2)} + \text{REP}(\bm{b}^{(2)})) \\ 
&= \bm{I}\circledast\bm{F}^{\prime} + \text{REP}(\bm{b}^\prime) \,.
\end{aligned}
\end{equation}
Transform IV is especially useful for generalizing Transform III to the groupwise case. Intuitively, a groupwise conv splits the input into $g$ parallel groups, convolves separately, then concatenates the outputs. To replace a $g$-group conv, we build a DBB where all the conv layers have the same groups $g$. For converting the $1\times1$ - $K\times K$ sequence, we equivalently split it into $g$ groups, perform Transform III separately, and concatenate the outputs (Fig. \ref{fig-dw}).

\paragraph{Transform V: a conv for average pooling} An average pooling with kernel size $K$ and stride $s$ applied to $C$ channels is equivalent to a conv with the same $K$ and $s$. Such a kernel $\bm{F}^\prime \in \mathbb{R}^{C\times C\times K\times K}$ is constructed by
\begin{equation}
\bm{F}^\prime_{d,c,:,:} = 
\begin{dcases}
\frac{1}{K^2} & \text{if $d=c$} \,, \\
0 & \text{elsewise} \,.
\end{dcases}
\end{equation}
Just like a common average pooling, it performs downsampling when $s>1$ but is actually smoothing when $s=1$.

\begin{figure}
	\begin{center}
		\includegraphics[width=0.9\linewidth]{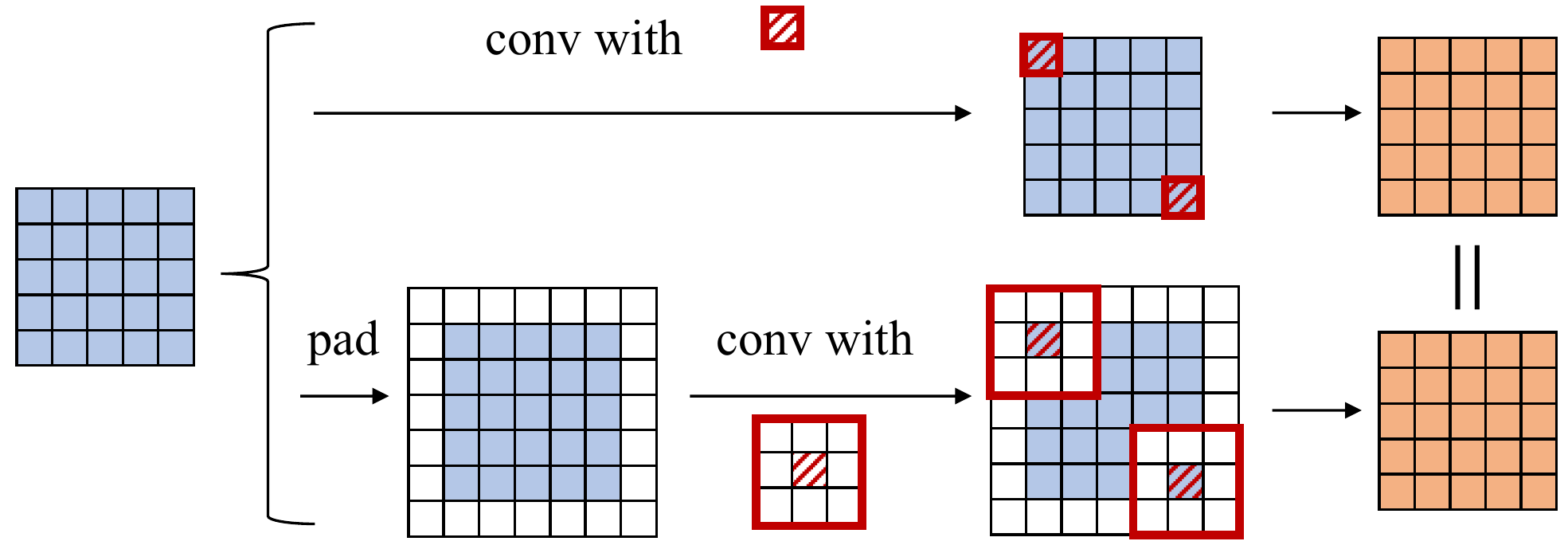}
		\vspace{-0.1in}
		\caption{An example of converting a $1\times1$ layer to $3\times3$ via Transform VI. To align the starting and ending points of sliding windows (shown at the top-left and bottom-right corners), the $3\times3$ layer should pad the input by one pixel.}
		\label{fig-pad-issue}
	\end{center}
	\vspace{-0.2in}
\end{figure}

\paragraph{Transform VI: a conv for multi-scale convolutions}
Considering a $k_h \times k_w$ ($k_h \leq K, k_w \leq K$) kernel is equivalent to a $K\times K$ kernel with some zero entries, we can transform a $k_h \times k_w$ kernel into $K\times K$ via zero-padding. Specifically, $1\times 1$, $1\times K$ and $K\times 1$ conv are particularly practical as they can be efficiently implemented. The input should be padded to align the sliding windows (Fig. \ref{fig-pad-issue}).

\subsection{An Inception-like DBB Instance}\label{sect-instance}
We present a representative instance of DBB (Fig. \ref{fig-topology}), while its universality and flexibility enable numerous feasible instances. Like Inception, we use $1\times1$, $1\times1$ - $K\times K$, $1\times1$ - AVG to enhance the original $K\times K$ layer. For the $1\times1$ - $K\times K$ branch, we set the internal channels equal to the input and initialize the $1\times1$ kernel as an identity matrix. The other conv kernels are initialized regularly \cite{initkaiming}. A BN follows every conv or AVG layer, which provides training-time nonlinearity. Without such nonlinearity, the performance gain will be marginal (Table. \ref{table-diversity}). Notably, for a depthwise DBB, every conv shall have the same number of groups, and we remove the $1\times1$ path and the $1\times1$ conv in the $1\times1$ - AVG path because $1\times1$ depthwise conv is just a linear scaling.

\setlength{\tabcolsep}{4pt}
\begin{table*}
	\begin{center}
		\caption{Experimental configurations.} 
		\label{table-configs}
		\small
		\vspace{-0.1in}
		\begin{tabular}{llcccccl}
			\toprule
			Dataset			&	Architecture	&	GPUs	&	Epochs / iterations	&	\makecell{Batch \\ size}	&	\makecell{Init \\ learn rate}		&	\makecell{Weight \\ decay} &	Data augmentation\\
			\midrule
			CIFAR-10/100		&	VGG-16&	1				&	600 epochs		&	128		&	0.1		&	$1\times10^{-4}$	&	crop + flip\\
			ImageNet		&	AlexNet			&	4		&	90	epochs		&	512		&	0.1		&	$5\times10^{-4}$	&	crop + flip\\
			ImageNet		&	MobileNet		&	8		&	90	epochs		&	256		&	0.1		&	$4\times10^{-5}$	&	crop + flip\\
			ImageNet		&	ResNet-18/50	&	8		&	120	epochs		&	256		&	0.1		&	$1\times10^{-4}$	&	+ color jitter + PCA lighting \cite{krizhevsky2012imagenet}	\\
			COCO detection	&	CenterNet \cite{centernet}	&	8		&	126k iters		&	128		&	0.02	&	$1\times10^{-4}$	&	+ color jitter + PCA lighting \cite{krizhevsky2012imagenet}	\\
			Cityscapes		&	PSPNet \cite{pspnet}			&	8		&	200 epochs		&	16		&	0.01	&	$1\times10^{-4}$	&	same as \cite{official-pspnet}	\\
			\bottomrule
		\end{tabular}
	\end{center}
\vspace{-0.15in}
\end{table*}
\setlength{\tabcolsep}{1.4pt}

\setlength{\tabcolsep}{4pt}
\begin{table*}
	\begin{center}
		\caption{Top-1 accuracy of the original model, ACNet \cite{ding2019acnet} and DBB-Net. The results on CIFAR are average of 5 runs}
		\label{table-improvements}
		\small
			\vspace{-0.1in}
		\begin{tabular}{llcccc}
			\toprule
			Dataset			&	Architecture	&	Original		&	ACNet 		&	DBB-Net 		&	Accuracy $\uparrow$\\
			\midrule
			CIFAR-10		&	VGG-16				&	93.95$\pm$0.03	&	94.43$\pm$0.03	&	\textbf{94.62}$\pm$0.02			&	0.67	\\
			CIFAR-100		&	VGG-16				&	74.05$\pm$0.10	&	75.30$\pm$0.04	&	\textbf{75.72}$\pm$0.07			&	1.67	\\
			\midrule
			ImageNet		&	AlexNet			&	57.23			&	58.43		&	\textbf{59.19}	&	1.96	\\
			ImageNet		&	MobileNet		&	71.89			&	72.14		&	\textbf{72.88}	&	0.99	\\
			ImageNet		&	ResNet-18		&	69.54			&	70.53		&	\textbf{70.99}	&	1.45	\\
			ImageNet		&	ResNet-50		&	76.14			&	76.46		&	\textbf{76.71}	&	0.57	\\
			\bottomrule
		\end{tabular}
	\end{center}
	\vspace{-0.25in}
\end{table*}
\setlength{\tabcolsep}{1.4pt}

\section{Experiments}

We use several benchmark architectures on CIFAR \cite{krizhevsky2009learning}, ImageNet \cite{deng2009imagenet}, Cityscapes \cite{cityscapes} and COCO detection \cite{coco} to evaluate the capability of DBB for improving ConvNet performance, and then investigate the significance of diverse connections and training-time nonlinearity.

\subsection{Datasets, Architectures and Configurations}

We first summarize the experimental configurations (Table. \ref{table-configs}). On \textbf{CIFAR-10/100}, we adopt the standard data augmentation techniques \cite{he2016deep}: padding to $40\times40$, random cropping and left-right flipping. We use VGG-16 \cite{simonyan2014very} for a quick sanity check. Following ACNet \cite{ding2019acnet}, we replace the two hidden fully-connected (FC) layers by global average pooling followed by one FC of 512 neurons. For the fair comparison, we equip each conv layer in the original models of VGG with BN. Then we use \textbf{ImageNet-1K}, which comprises 1.28M images for training and 50K for validation. For the data augmentation, we employ the standard pipeline including random cropping, left-right flipping for small models like AlexNet \cite{krizhevsky2012imagenet} and MobileNet \cite{mobilev1}, and additional color jitter and a PCA-based lighting for ResNet-18/50 \cite{he2016deep}. Specifically, we use the same AlexNet as ACNet \cite{ding2019acnet}, which is composed of five stacked conv layers followed by three FC layers with no local response normalizations. We insert BN after each conv layer as well. For the simplicity, we use cosine learning rate decay on CIFAR and ImageNet with an initial value of 0.1. On \textbf{COCO detection}, we train CenterNet \cite{centernet} in 126k iterations with a learning rate initialized as 0.02 and multiplied by 0.1 at the 81k and 108k iterations respectively. On \textbf{Cityscapes}, we simply adopt the official implementation and default configurations \cite{official-pspnet} of PSPNet \cite{pspnet} for the better reproducibility: poly learning rate with base of 0.01 and power of 0.9 for 200 epochs. 

For each architecture, we replace every $K\times K \,(1<K<7)$ conv and its following BN by a DBB to construct a DBB-Net. We do not experiment with larger kernels (\eg, the first $7\times7$ and $11\times11$ conv of ResNet and AlexNet) because they are less favored in model architectures. All the models are trained with identical configurations. After training, the DBB-Nets are converted into the same structure as the original model and tested. All the experiments are accomplished with PyTorch.

\subsection{DBB for Free Improvements}

Table. \ref{table-improvements} shows that the DBB-Nets exhibit a clear and consistent boost of performance on CIFAR and ImageNet: DBB improves VGG-16 on CIFAR-10 and CIFAR-100 by 0.67\% and 1.67\%, AlexNet on ImageNet by 1.96\%, MobileNet by 0.99\%, and ResNet-18/50 by 1.45\%/0.57\%, respectively. Even though ACB \cite{ding2019acnet} (Fig. \ref{fig-counterparts}d) is a special case of DBB, we still choose it as a competitor to compare with. Concretely, we add $K\times 1$ and $1\times K$ branches to construct ACBs, and train with the same settings. The superiority of DBB-Net over ACNet suggests that combining paths with Inception-like different complexities may benefit the model more than aggregating features generated by multi-scale convolutions. Notably, the comparisons are biased towards the original models, as we adopt the hyper-parameters reported in the original papers (\eg, weight decay of $10^{-4}$ on ResNets), which have been tuned on the original models but may be less suitable for the DBB-Nets.

\begin{figure*}
	\begin{subfigure}{0.33\linewidth}
		\includegraphics[width=\linewidth]{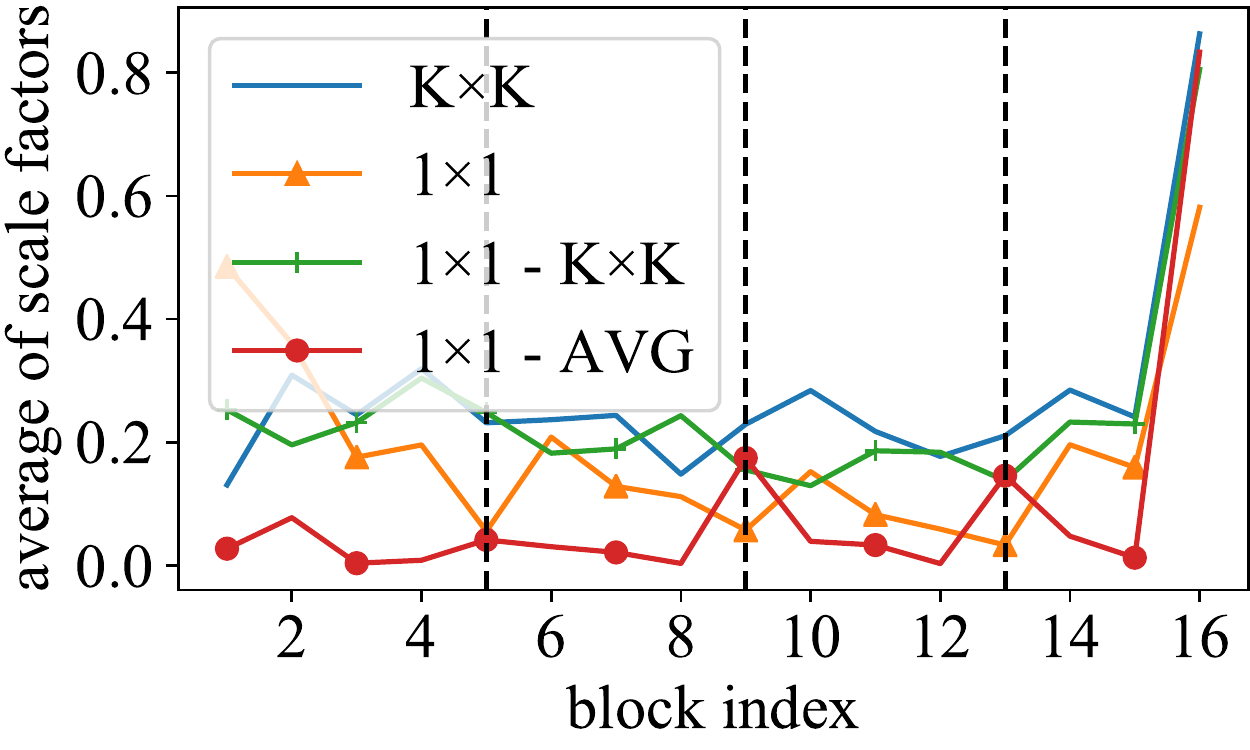} 
		\subcaption{Average magnitude of scale factors.}
		\label{fig-bn-across-layers}
	\end{subfigure}
	\begin{subfigure}{0.33\linewidth}
		\includegraphics[width=\linewidth]{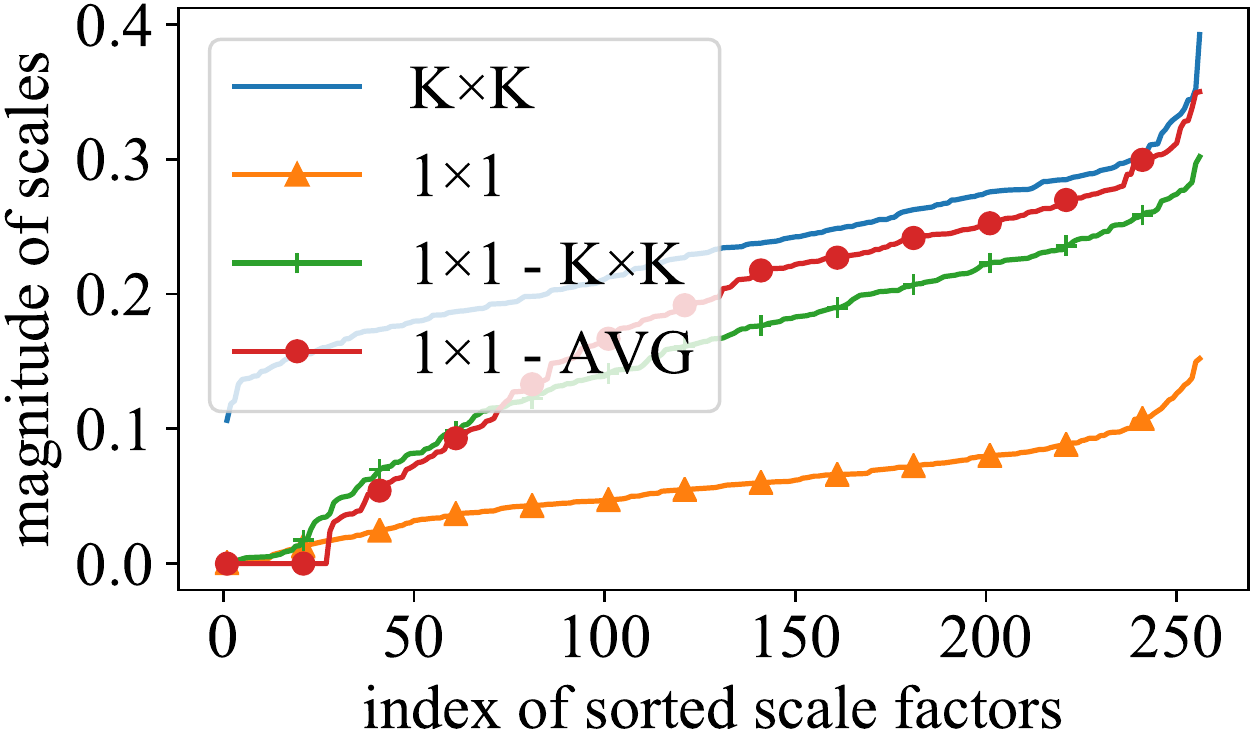}
		\subcaption{Sorted scale factors of stride-2 layer.}
		\label{fig-bn-block9}
	\end{subfigure}
	\begin{subfigure}{0.33\linewidth}
		\includegraphics[width=\linewidth]{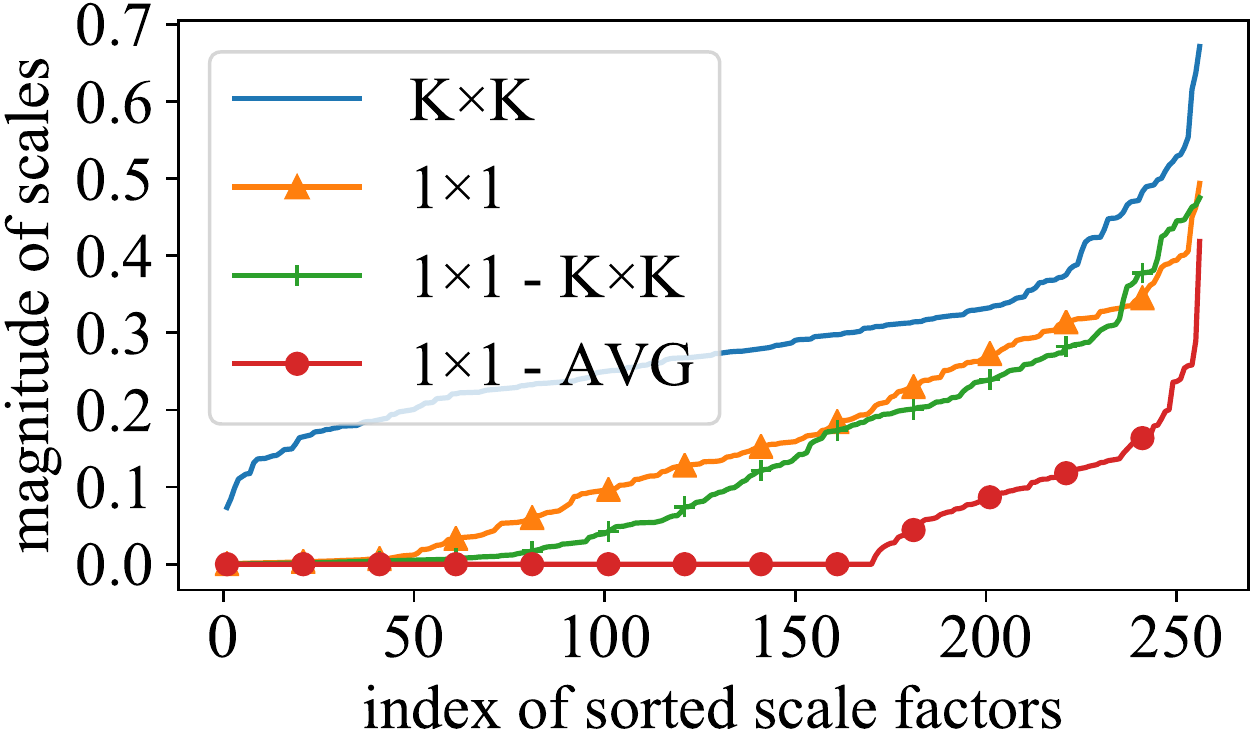} 
		\subcaption{Sorted scale factors of stride-1 layer.}
		\label{fig-bn-block10}
	\end{subfigure}
	\caption{Left: the average magnitude of scaling factors of BN in every DBB across different layers. Vertical dashed lines indicate the stage transition with stride-2 DBB. Middle and right: the magnitude of scaling factors sorted in ascending order of the 9th DBB (stride-2) and the 10th DBB (stride-1).}
	\label{fig-bn}
\end{figure*}

We continue to verify the significance of every branch by showing the scaling factors $\bm{\gamma}$ of the four BN layers before the addition. Specifically, for each of the 16 $3\times3$ DBBs of ResNet-18 (because it originally has 16 $3\times3$ conv layers), we compute the average of absolute value of the four scaling vectors. Table. \ref{fig-bn-across-layers} shows that the $K\times K$, $1\times1$ and $1\times1-K\times K$ branches have comparable magnitude of scaling factors, suggesting that the three branches are important. An interesting discovery is that the $1\times1$ - AVG branch is more important for a stride-2 DBB, suggesting the average pooling is more useful as downsampling than smoothing. Fig. \ref{fig-bn-block9} shows the absolute values of scaling factors of the 9th block with the four $\bm{\gamma}$ vectors respectively sorted for the better readability. It is observed that the $K\times K$ branch have a larger minimum scale, and the scales of the other three branches have a wide range. The phenomenon is quite different for the 10th block (Fig. \ref{fig-bn-block10}), which has stride=1: the scales of $1\times1$ - AVG branch are close to zero for more than 160 channels but relatively large for the others, and the scales of $1\times1$ branch are larger than the 9th block, which suggests that $1\times1$ conv is more useful with stride=1. Such a diversity of distributions of scaling factors suggest that the DBB-Net learns a diverse combination of the diverse branches for each block, and the discoveries may shed light on other research areas like architecture design.


\subsection{Object Detection and Semantic Segmentation}

We use the ImageNet-pretrained ResNet-18 models to verify the generalization performance on object detection and semantic segmentation. Specifically, we build two CenterNets/PSPNets where the only difference is the backbone (the original ResNet-18 or DBB-ResNet-18), load the ImageNet-pretrained (not yet transformed) weights, train on COCO/Cityscapes, perform the transformations and test.  
\setlength{\tabcolsep}{4pt}
\begin{table}
	\begin{center}
		\caption{Object detection and semantic segmentation.}
		\label{table-det-seg}
		\small
		\begin{tabular}{lccc}
			\toprule
			Backbone			&ImageNet top-1	&	COCO AP		&	Cityscapes mIoU		\\
			\midrule
			Original Res18		&	69.54		&	29.83		&	70.18	\\
			DBB-Res18			&	70.99		&	30.68		&	71.35\\
			\bottomrule
		\end{tabular}
	\end{center}
	\vskip -0.2in
\end{table}
\setlength{\tabcolsep}{1.4pt}

\setlength{\tabcolsep}{4pt}
\begin{table*}
	\caption{Top-1 accuracy of ResNet-18 on ImageNet with different blocks. The training speed (batches/second) is recorded on the same machine with eight 1080Ti GPUs. The training-time eval speed (batches/s) is tested with the original (\ie, not yet transformed) model on a single GPU with a batch size of 128. For reference, the parameters and speed of \textit{every inference-time model} (because all the models end up with the same inference-time structure) are 11.68M and 19.95 batches/s.}
	\vskip -0.2in
	\label{table-diversity}
	\begin{center}
		\small
		\begin{tabular}{l|l|c|c|c|c|c|c|c|c}
			\toprule
			&Block	&		\makecell{Original \\ $K\times K$} &	\makecell{$1\times1$}	&	\makecell{$1\times1$ -\\ $K\times K$} &	\makecell{$1\times1$ -\\ AVG}	& Accuracy & \makecell{Training \\ param (M)}	& \makecell{Training \\ speed} & \makecell{Training-time \\ eval speed} \\
			\midrule
			\multirow{11}{*}{With BN}	&	DBB	& 	1	&\checkmark	&\checkmark	&\checkmark	&70.99		&26.33	&	4.06	&	4.11\\
			& 	DBB	&	1	&			&\checkmark	&\checkmark &70.36		&25.09						&	4.16	&	4.30\\
			& 	DBB	&	1	&\checkmark	&			&\checkmark &70.40		&14.18						&	4.31	& 	6.64\\													
			& 	DBB	&	1	&\checkmark	&\checkmark	&			&70.74		&25.08						&	4.21	&	6.33\\
			&	DBB	&	1	&\checkmark	&			&			&70.15		&12.93						&	4.38	&	14.2\\
			& 	DBB	&	1	&			&\checkmark	&			&70.20		&23.84						&	4.22	&	7.31\\
			& 	DBB	&	1	&			&			&\checkmark	&70.02		&12.95						&	4.33	&	7.59\\
			&	Baseline	&	1	&			&			&	&69.54		&11.69						&	4.44	&	19.24\\
			&	Baseline + init &	1&	&					&	&69.67		&11.69						&	4.44	&	19.24\\
			&	Double Duplicate	&	2	&		&	&		&69.81		&22.69						&	4.36	&	11.04\\
			&	Triple Duplicate	&	3	&		&	&		&70.29		&33.70						&	4.20	&	7.75\\
			\midrule
			\multirow{3}{*}{Purely Linear}&	DBB	&	1	&\checkmark	&\checkmark	&\checkmark	&70.12		&26.20	&	-	&	-\\	
			&	DBB	&	1	&\checkmark	&			&			&69.83		&12.91&	-	&	-\\	
			&	Double Duplicate&	2&	&			&			&69.59		&22.68&	-	&	-\\	
			\bottomrule
		\end{tabular}
	\end{center}
	\vskip -0.1in
\end{table*}
\setlength{\tabcolsep}{1.4pt}

\subsection{Ablation Studies}\label{sect-diversity}
\begin{figure*}
	\begin{center}
		\includegraphics[width=0.9\linewidth]{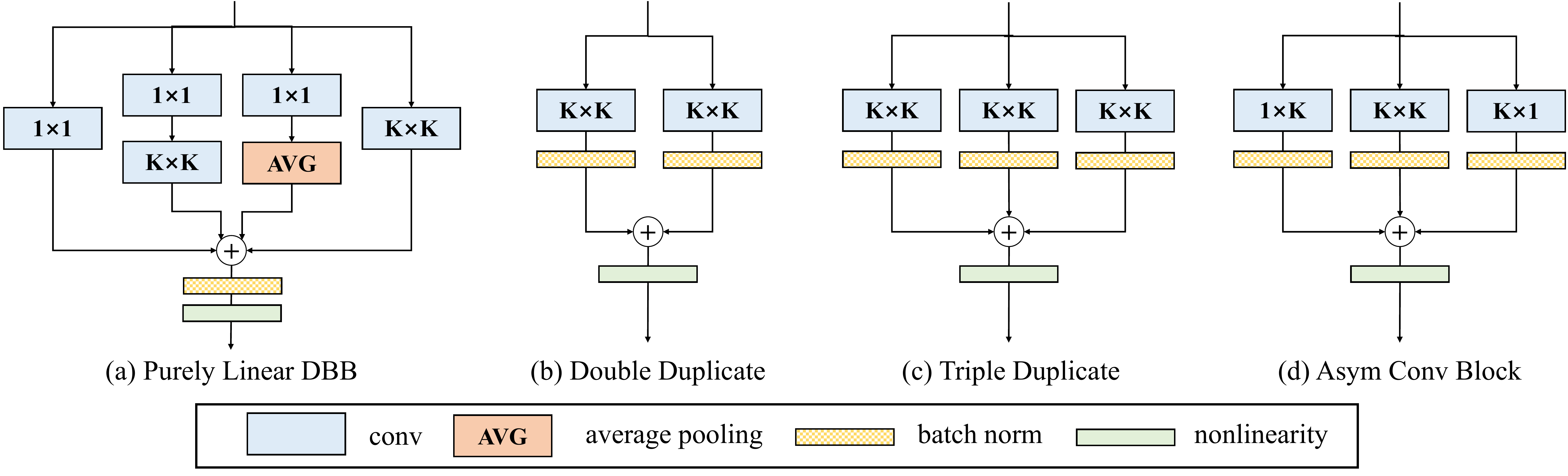}
		\caption{Counterparts to compare against DBB.}
		\label{fig-counterparts}
	\end{center}
\end{figure*}

We conduct a series of ablation studies on ResNet-18 to verify the significance of diverse connections and training-time nonlinearity. Specifically, we first ablate some branches from DBB and observe the change in performance, then compare DBB to some counterparts with duplicate branches or purely linear combination of branches, as shown in Fig. \ref{fig-counterparts}. For the purely linear counterpart, we use no BN before the branch addition, but the sum passes through BN. Again, all the models are trained from scratch with the same settings as before and converted into the same original structure for testing. We present in Table. \ref{table-diversity} the final accuracy and the training costs.

Table. \ref{table-diversity} shows that removing any branch degrades the performance, suggesting that every branch matters. It is also observed that using any of the three branches can lift the accuracy to above 70\%. Seen from the training-time parameters \vs accuracy, one may use a lightweight DBB with only the $1\times1$ and $1\times1$ - AVG branches for lower accuracy but more efficient training, if the training resources are limited. The Double/Triple Duplicate blocks also improve the accuracy, but not as much as diverse branches do. We have two especially interesting discoveries when comparing DBB to duplicate blocks with the same number of branches: 

\begin{itemize}[noitemsep,nolistsep,topsep=0pt,parsep=0pt,partopsep=0pt]
	\item A $1\times1$ conv can be viewed as a degraded $3\times3$ conv with many zero entries, which has weaker representational capacity than the latter, but the accuracy is 70.15\% for ($K\times K$ + $1\times1$) and 69.81\% for double $K\times K$. In other words, a weak-capacity component plus a strong-capacity one is better than two strong components.
	\item Similarly, the DBB with ($K\times K$ + $1\times 1$ + ($1\times1$ - AVG)) outperforms triple $K\times K$ (70.40\% \textgreater 70.29\%), though the latter has 2.3$\times$ training-time parameters as the former, suggesting that the representational capacity of ConvNet is determined by not only the amount of parameters but also the diversity of connections.
\end{itemize}

To verify if the improvements are due to the different initialization, we construct a baseline (denoted by ``baseline + init'') by transforming the full-featured DBB-Net right after random initialization, using the resultant weights to initialize a regular ResNet-18, and then training it with the same settings. The final accuracy is 69.67\%, which is hardly higher than the baseline with regular initialization, suggesting that initialization is not the key.

We continue to validate the training-time nonlinearity brought by the BN in branches. In the above discussions, we have noticed that even duplicate branches with BN can improve the performance, as such training-time nonlinearity makes the block more powerful than a single conv. When the BN layers are moved from pre- to post-addition, the block (from the input to the branch addition) becomes purely linear during training. In this case, the Double Duplicate block hardly improves the performance (69.54\% $\to$ 69.59\%), and the DBB of ($K\times K$ + $1\times1$) improves not as much as the comparable DBB with BN (69.83\% $<$ 70.15\%), suggesting that diverse connections can improve the model even without training-time nonlinearity.

We also present the training speed and the inference speed of the training-time models in Table. \ref{table-diversity}, which shows that increasing the training-time parameters does not significantly slow down the training speed. Notably, the actual training speed is influenced by the data preprocessing, cross-GPU communication, implementation of backpropagation, \etc, hence such data are for reference only. In industry, the researchers and engineers usually have abundant training resources but strict restrictions on the inference-time costs, so they may intend to train the models for tens of extra days for very minor performance improvements. In these application scenarios, one may find DBB particularly useful for building powerful ConvNets with only reasonable extra training costs.

\section{Conclusions}
We proposed a ConvNet building block named DBB, which implements the combination of diverse branches via a single convolution. DBB allows us to improve the performance of off-the-shelf ConvNet architectures with absolutely no extra inference-time costs. Through controlled experiments, we demonstrated the significance of diverse connections and training-time nonlinearity, which make a DBB more powerful than a regular conv layer, though they end up with the same inference-time structure.

{\small
\bibliographystyle{ieee_fullname}
\bibliography{dbbbib}
}

\end{document}